\documentclass{article}

    \PassOptionsToPackage{numbers, compress}{natbib}

\usepackage[preprint]{neurips_2026}

\usepackage[utf8]{inputenc} 
\usepackage[T1]{fontenc}    
\usepackage{hyperref}
\usepackage{url}            
\usepackage{graphicx}
\usepackage{booktabs}       
\usepackage{algorithm}
\usepackage{algpseudocode}
\usepackage{amsmath}
\usepackage{amsthm}
\usepackage{amsfonts}       
\usepackage{nicefrac}       
\usepackage{microtype}      
\usepackage{xcolor}         
\usepackage{multirow}
\usepackage{wrapfig}
\usepackage{tabularx}
\usepackage[table]{xcolor}

\definecolor{globalsym}{RGB}{33,102,140}
\definecolor{headsym}{RGB}{173,94,0}
\definecolor{ffnsym}{RGB}{156,54,54}

\title{Scaling Linear Mode Connectivity and Merging to Billion Parameter Pretrained Transformers}

\author{%
    Tianyi Li\\
  MBZUAI\\
  \texttt{Tianyi.Li@mbzuai.ac.ae} \\
  \And
  Zhiqiang Shen \\
  MBZUAI \\
  \texttt{Zhiqiang.Shen@mbzuai.ac.ae} \\
}

\begin{document}
\maketitle

\begin{abstract}

Linear mode connectivity (LMC) provides a promising foundation for understanding and merging independently trained neural networks, but existing methods typically optimize the interpolation path from only one model endpoint, limiting their scalability and effectiveness for large pretrained transformers. We propose a novel and scalable framework for enabling LMC-based model merging to {\em billion-parameter pretrained transformers}. Our method applies properly parameterized functionality-preserving weight transformations to align functionally equivalent solutions, and introduces a dual learning procedure in which both models jointly learn their corresponding transformations toward a shared linear interpolation path. This bidirectional optimization substantially reduces interpolation barriers and enables more reliable merging across large-scale architectures. Empirically, we show that our approach achieves near-zero loss barriers on WikiText for language models with medium-sized parameters, representing, to our knowledge, the first demonstration of near-barrier-free linear connectivity at this scale. In the vision domain, ViT-L maintains above 69\% ImageNet top-1 accuracy throughout the interpolation path, while modern billion-parameter LLMs exhibit only small loss barriers. These results suggest that properly resolving parameter symmetries enables large pretrained Transformers to be connected and merged through simple linear paths with substantially improved interpolation performance. 
Code: \url{https://github.com/VILA-Lab/Dual-Learned-Matching}.

\end{abstract}

\section{Introduction}

Deep neural network weights have recently become an increasingly important object of study in the machine learning community~\cite{han2026survey,garipov2018loss,ainsworth2022git,zhao2025symmetry,frankle2020linear}. Beyond specifying a model's input-output behavior, trained weights are increasingly viewed as reusable artifacts that can be inspected, edited, adapted, merged, or even treated as data for training other models~\cite{wang2024neural,zeng2025generative}. This perspective is especially important in the era of large pretrained models, where each checkpoint represents substantial investment in data, computation, optimization, and engineering~\cite{minaee2024large}. Understanding the structure of weight space, and how different checkpoints are geometrically and functionally related, is therefore not only a theoretical question about loss landscapes, but also a practical foundation for scalable model reuse and composition.

A central idea for studying weight-space structure is linear mode connectivity (LMC)~\cite{garipov2018loss,draxler2018essentially}. Two models are linearly mode connected if the linear interpolation between their weights remains in a low-loss region. When such connectivity exists, simple weight interpolation becomes a natural and efficient mechanism for model merging, without requiring ensembles, additional architectures, or expensive retraining. However, independently trained models often fail to exhibit low-loss linear paths in their raw parameterization. Even when two networks implement similar functions, they may occupy apparently distant regions of weight space due to differences in initialization, training order, datasets, and hyperparameters~\cite{frankle2020linear,kwok2025butterfly}. As a result, naive interpolation can encounter large loss barriers, making model merging unreliable.

A major reason for this failure is that neural networks exhibit many function-preserving symmetries: distinct parameter configurations can represent the same underlying function~\cite{brea2019weight,entezari2021role}. A classic example is neuron permutation symmetry, where hidden units within a layer can be permuted together with corresponding inverse permutations in adjacent layers without changing the network function~\cite{hecht1990algebraic,chen1993geometry,entezari2021role}. In small networks, resolving such symmetries can align independently trained models before interpolation and reveal low-loss paths that are hidden in the original coordinates~\cite{entezari2021role,singh2020model,tatro2020optimizing}. Existing alignment methods commonly rely on activation matching, which uses data to align intermediate representations~\cite{li2015convergent,tatro2020optimizing,ainsworth2022git}, or weight matching, which aligns parameters directly in a data-free manner~\cite{singh2020model,ainsworth2022git}.

Beyond discrete symmetries like permutation, Transformers, however, introduce a substantially richer and more structured symmetry space than standard multilayer networks. Their residual connections, normalization layers, multi-head attention modules, positional encodings, and feed-forward blocks induce multiple classes of functionality-preserving transformations~\cite{ashkboos2024slicegpt,theus2025generalized}. Recent studies have begun to exploit these Transformer-specific symmetries to uncover hidden connectivity between trained models~\cite{zhang2025beyond}. Nevertheless, most activation- or weight-matching methods remain task-agnostic: they align representations or parameters without directly optimizing the performance of interpolated models. This mismatch limits their ability to reduce the actual loss barrier, especially for pretrained Transformers where small misalignments can accumulate across many layers and severely degrade interpolation performance.

\begin{figure}[t]
\centering
\setlength{\textfloatsep}{6pt plus 1pt minus 2pt}
\vspace{-0.7em}
\setlength{\abovecaptionskip}{2pt}
\setlength{\belowcaptionskip}{-6pt}
\includegraphics[width=0.98\linewidth]{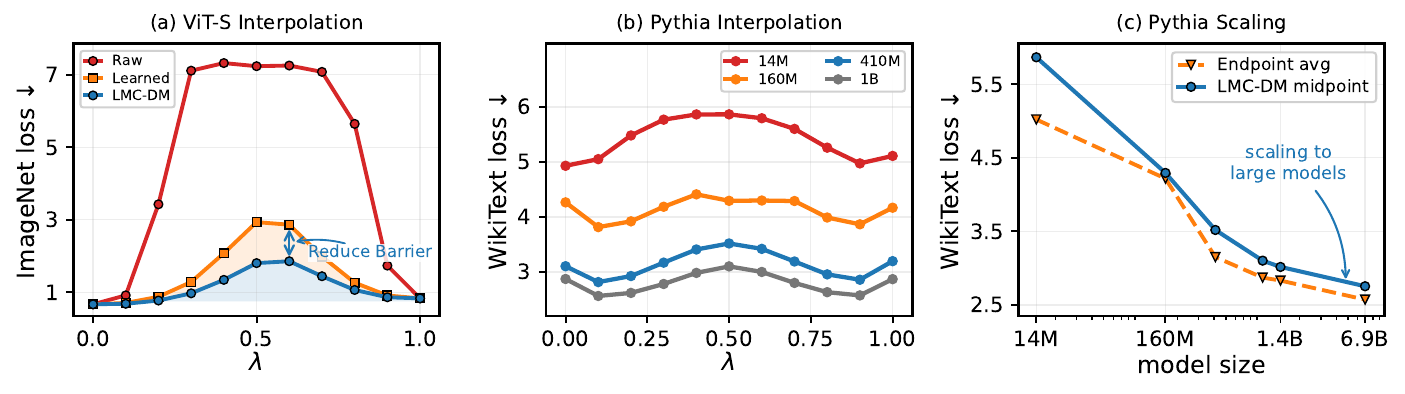}
\vspace{-0.45em}
\caption{Loss barriers and scaling behavior.
Left: Interpolation loss barriers for ViT-S under naive weight interpolation, learned matching, and LMC-DM (Ours). Middle: Interpolation loss curves for Pythia models across different parameter scales. Right: Scaling behavior of LMC-DM–merged Pythia models compared to endpoint models as model size increases.}
\label{fig:intro-curves}
\vspace{-1.em}
\end{figure}

Learned matching partially addresses this issue by optimizing symmetry transformations using the task loss of interpolated models~\cite{theus2025generalized}. However, existing learned matching approaches still have several limitations. First, they typically exploit only a subset of the full Transformer symmetry group and often rely on simple parameterizations, such as unconstrained variables that are projected during the forward pass. Second, they apply symmetry transformations to only one endpoint model, leading to an asymmetric formulation in which one model is optimized toward a fixed counterpart. This one-sided optimization can create suboptimal landscapes and restrict the search for better-aligned representatives of both models. Consequently, prior demonstrations have largely been limited to models with millions or tens of millions of parameters trained on relatively small datasets. Whether LMC can be effectively revealed in large pretrained Transformers, and whether symmetry learning can scale to billion-parameter models, remains largely unexplored.

In this work, we propose a scalable framework for linear mode connectivity and model merging in billion-parameter pretrained Transformers. We systematically formulate a broad family of functionality-preserving weight transformations for Transformers, including normalization absorption, residual-space rotation and scaling, attention-head permutation, head-internal transformations, and feed-forward permutation and scaling. We parameterize these transformations under appropriate structural constraints and optimize them directly with respect to the loss along the interpolation path. Most importantly, we introduce a dual learning procedure, where both endpoint models learn their own symmetry transformations toward a shared linear path. This bidirectional formulation gives both models the flexibility to resolve their internal symmetries, substantially reduces interpolation barriers, and enables more reliable merging at scale. Empirically, we show that ViT-L maintains over 69\% ImageNet-1K top-1 accuracy across the entire interpolation path; language models on 160M parameters achieve near-zero loss barriers on WikiText, the first such result in this setting, outperforming all prior counterparts such as 
Weight Matching~\cite{ainsworth2022git} and Learned Matching~\cite{theus2025generalized};
and modern billion-parameter LLMs exhibit only small loss barriers. These results suggest that large pretrained Transformers can be effectively aligned and merged by simple linear interpolation when equipped with properly parameterized and jointly optimized symmetry transformations.

Our contributions are summarized as follows:

\begin{itemize}
    \item We formulate a broad family of functionality-preserving Transformer symmetries. Specifically, our framework covers normalization absorption, residual-space rotation and scaling, attention-head permutation, head-internal transformations, and feed-forward permutation and scaling. We investigate practical parameterizations of continuous symmetry variables, enabling functionality-preserving transformations to be learned directly from the interpolation loss.
    \item We propose a dual learning procedure for bidirectional endpoint alignment. Unlike prior one-sided methods, both models learn their own transformations toward a shared linear interpolation path, leading to better-conditioned optimization and lower interpolation barriers.
    \item We show strong and extensive empirical results that LMC scales to large pretrained Transformers. We demonstrate near-zero loss barriers for language models with 160M parameters on WikiText, over 69\% ImageNet top-1 accuracy along the interpolation path for ViT-L, and only small loss barriers for billion-parameter LLMs.
\end{itemize}
\section{Functionality-Preserving Symmetries in Transformers}
\label{sec:pre}

\begin{figure}[t]
\centering
\setlength{\abovecaptionskip}{2pt}
\setlength{\belowcaptionskip}{-4pt}
\includegraphics[width=0.9\linewidth]{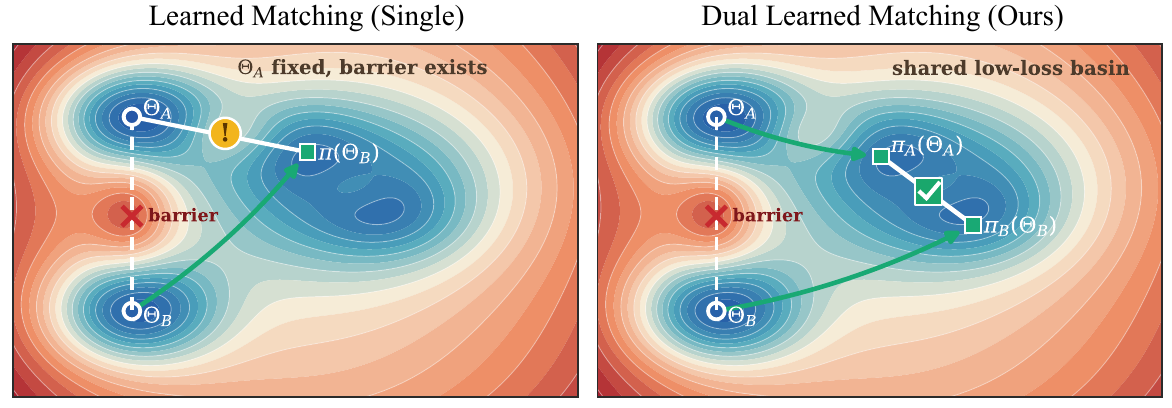}
\vspace{-4pt}
\caption{Schematic loss-landscape view of single and dual learned matching. Direct interpolation crosses a high-loss barrier. Single learned matching reduces but does not eliminate this barrier by optimizing one checkpoint toward a fixed reference. Dual learned matching jointly optimizes both checkpoints, yielding a lower barrier in a shared low-loss basin.}
\label{fig:single_dual_landscape}
\end{figure}

\newcommand{\symbox}[2]{\begingroup\setlength{\fboxsep}{1pt}\colorbox{#1!16}{$#2$}\endgroup}

This section introduces the functionality-preserving symmetry family used in our work. A symmetry is a reparameterization of the weights that changes the coordinates of internal representations while leaving the network function unchanged. Our goal is to expose equivalent parameterizations of two pretrained Transformers so that their weights can be better aligned before linear interpolation. In particular, we first convert normalization layers into parameter-free RMSNorm operators, which simplifies subsequent merging and enables global orthogonal symmetries on the residual stream. We then describe residual-space, attention-head, attention-circuit, and FFN symmetries. The highlighted symbols in the equations denote symmetry parameters.

\subsection{Normalization and Residual-Space Symmetry}

\noindent{\bf Reparameterization of Normalization Layers.}
Pretrained Transformers usually use normalization layers with learnable affine parameters, whereas the symmetry above is stated for parameter-free RMSNorm. 
Previously, \citet{ashkboos2024slicegpt} showed that LayerNorm and RMSNorm with parameters can be exactly converted into parameter-free RMSNorm by absorbing the affine parameters into adjacent linear layers.
Given an RMSNorm layer with scaling gain $\boldsymbol{\gamma}$ followed by a linear map $\mathbf{W}$,
\begin{equation}
    \mathrm{RMSNorm}_{\boldsymbol{\gamma}}(\mathbf{X})\mathbf{W}  = \mathrm{RMSNorm}_{0}(\mathbf{X})\,
    \mathrm{Diag}(\boldsymbol{\gamma})\mathbf{W},
\end{equation}
where $\mathrm{RMSNorm}_{0}$ denotes RMSNorm without affine parameters. Thus the gain can be exactly absorbed into the following linear layer, allowing us to treat RMSNorm as parameter-free in the symmetry analysis.
For pretrained Transformers that use LayerNorm, the conversion requires an additional centering step. Let
\begin{equation}
    \mathbf{C} = \mathbf{I} - \frac{1}{d}\mathbf{1}\mathbf{1}^{\top}
\end{equation}
be the channel-centering operator. Writing $\mathrm{LayerNorm}_{0}$ for LayerNorm without affine parameters, we have
\begin{equation}
    \mathrm{LayerNorm}_{0}(\mathbf{X})
    =
    \mathrm{RMSNorm}_{0}(\mathbf{X}\mathbf{C}).
\end{equation}
With affine parameters followed by a linear map,
\begin{equation}
    \mathrm{LayerNorm}_{\boldsymbol{\gamma},\boldsymbol{\beta}}(\mathbf{X})\mathbf{W}
    =
    \mathrm{RMSNorm}_{0}(\mathbf{X}\mathbf{C})
    \mathrm{Diag}(\boldsymbol{\gamma})\mathbf{W}
    +
    \boldsymbol{\beta}\mathbf{W}.
\end{equation}
Thus the LayerNorm gain is absorbed into the following weight, and the LayerNorm bias is absorbed into the following bias term. By projecting residual writes onto the zero-mean subspace, a pre-LayerNorm Transformer can be represented as a parameter-free pre-RMSNorm Transformer on the centered residual stream. Appendix~\ref{app:norm-details} gives the exact weight and bias transformations.

\noindent{\bf Global Residual-Space Symmetry.}
A Transformer using parameter-free RMSNorm has a global symmetry in the residual coordinate system. Under a row-vector convention, write one layer as
\begin{equation}
\mathbf{X}_{\ell+1}
=
\mathbf{X}_\ell
+
F_\ell(\mathrm{RMSNorm}(\mathbf{X}_\ell)).
\end{equation}
Consider a scalar $\symbox{globalsym}{g}>0$ and an orthogonal matrix $\symbox{globalsym}{\mathbf{G}}$, shared by all layers, that transform the residual stream as $\mathbf{X}_\ell'=\mathbf{X}_\ell\symbox{globalsym}{g\mathbf{G}}$. RMSNorm removes the scalar but preserves the rotation:
\begin{equation}
\mathrm{RMSNorm}(\mathbf{X}\,\symbox{globalsym}{g\mathbf{G}})
=
\mathrm{RMSNorm}(\mathbf{X})\,\symbox{globalsym}{\mathbf{G}},
\end{equation}
so each residual-read linear map can absorb the rotation by multiplying its input weight by $\symbox{globalsym}{\mathbf{G}}^\top$. To keep the residual stream in the transformed coordinates, every residual write, including attention output projections and FFN down projections, is also multiplied by $\symbox{globalsym}{g\mathbf{G}}$. Denote the block after these weight transformations by $F_\ell'$. It writes the original update in the new residual coordinates, giving
\begin{equation}
\mathbf{X}_{\ell+1}'
=
\mathbf{X}_\ell'
+
F_\ell'(\mathrm{RMSNorm}(\mathbf{X}_\ell'))
=
\left(
\mathbf{X}_\ell
+
F_\ell(\mathrm{RMSNorm}(\mathbf{X}_\ell))
\right)
\symbox{globalsym}{g\mathbf{G}}
=
\mathbf{X}_{\ell+1}\symbox{globalsym}{g\mathbf{G}} .
\end{equation}
Therefore, $\symbox{globalsym}{\mathbf{G}}$ can be viewed as a global coordinate freedom of the residual stream.

\subsection{Attention Head Permutations}

Multi-head attention is invariant to permutations of its heads. For an input $\mathbf{X}$, write one attention layer as
\begin{equation}
\mathrm{MHA}(\mathbf{X})
=
\sum_{i=1}^{h}
\mathrm{softmax}
\left(
\frac{\mathbf{X}\mathbf{W}_{Q,i}
(\mathbf{X}\mathbf{W}_{K,i})^\top}{\sqrt{d_h}}
\right)
\mathbf{X}\mathbf{W}_{V,i}\mathbf{W}_{O,i}.
\end{equation}
For any head permutation $\pi$, let $\symbox{headsym}{\mathbf{P}_h}=\symbox{headsym}{\mathbf{P}}\otimes\mathbf{I}_{d_h}$ be the corresponding lifted permutation over concatenated head channels. In per-head notation, the reparameterization
\begin{equation}
\mathbf{W}_{Q,i}'=\mathbf{W}_{Q,\pi(i)},\quad
\mathbf{W}_{K,i}'=\mathbf{W}_{K,\pi(i)},\quad
\mathbf{W}_{V,i}'=\mathbf{W}_{V,\pi(i)},\quad
\mathbf{W}_{O,i}'=\mathbf{W}_{O,\pi(i)}
\end{equation}
only reorders the summands, and therefore
\begin{equation}
\mathrm{MHA}'(\mathbf{X})=\mathrm{MHA}(\mathbf{X}).
\end{equation}

\subsection{QK and OV Circuit Symmetries}

Each attention head admits internal symmetries that preserve the attention logits and output.

For the QK circuit, multiplying the query projection by any invertible matrix can be compensated by multiplying the key projection by its inverse transpose. Concretely, for any invertible matrix $\symbox{headsym}{\mathbf{M}_{Q,i}}$,
\begin{equation}
    \mathbf{W}_{Q,i}'=\mathbf{W}_{Q,i}\symbox{headsym}{\mathbf{M}_{Q,i}},
    \qquad
    \mathbf{W}_{K,i}'=\mathbf{W}_{K,i}\symbox{headsym}{\mathbf{M}_{Q,i}}^{-\top},
\end{equation}
which preserves the attention logits:
\begin{equation}
    \mathbf{X}\mathbf{W}_{Q,i}'
    (\mathbf{X}\mathbf{W}_{K,i}')^{\top}
    =
    \mathbf{X}\mathbf{W}_{Q,i}
    (\mathbf{X}\mathbf{W}_{K,i})^{\top}.
\end{equation}

For the OV circuit, multiplying the value projection by any invertible matrix can be compensated by multiplying the output projection by its inverse. For any invertible matrix $\symbox{headsym}{\mathbf{M}_{V,i}}$,
\begin{equation}
    \mathbf{W}_{V,i}'=\mathbf{W}_{V,i}\symbox{headsym}{\mathbf{M}_{V,i}},
    \qquad
    \mathbf{W}_{O,i}'=\symbox{headsym}{\mathbf{M}_{V,i}}^{-1}\mathbf{W}_{O,i},
\end{equation}
which preserves the value-output product:
\begin{equation}
    \mathbf{X}\mathbf{W}_{V,i}'\mathbf{W}_{O,i}'
    =
    \mathbf{X}\mathbf{W}_{V,i}\mathbf{W}_{O,i}.
\end{equation}

In practice, additional architectural constraints (e.g., positional encodings such as RoPE~\cite{su2024roformer}) can further restrict these symmetries.

\subsection{FFN Permutation}

A standard two-layer FFN admits hidden-channel permutations. For
\begin{equation}
\mathrm{FFN}(\mathbf{X})
=
\phi(\mathbf{X}\mathbf{W}_{1}+\mathbf{b}_1)\mathbf{W}_{2}
+\mathbf{b}_2,
\end{equation}
any hidden permutation $\symbox{ffnsym}{\mathbf{P}_f}$ yields
\begin{equation}
\mathbf{W}_{1}'=\mathbf{W}_{1}\symbox{ffnsym}{\mathbf{P}_f},\qquad
\mathbf{b}_{1}'=\mathbf{b}_{1}\symbox{ffnsym}{\mathbf{P}_f},\qquad
\mathbf{W}_{2}'=\symbox{ffnsym}{\mathbf{P}_f}^{\top}\mathbf{W}_{2},
\end{equation}
and therefore
\begin{equation}
\mathrm{FFN}'(\mathbf{X})=\mathrm{FFN}(\mathbf{X}).
\end{equation}

\paragraph{GLU rescaling symmetry.}
Beyond the hidden-channel permutation above, GLU-style FFNs have an additional value-branch rescaling symmetry: scaling the value branch can be canceled exactly in the down projection, while the gate branch is unchanged. For
\begin{equation}
\mathrm{GLU}(\mathbf{X})
=
\left(
\phi(\mathbf{X}\mathbf{W}_{\mathrm{gate}})
\odot
\mathbf{X}\mathbf{W}_{\mathrm{up}}
\right)
\mathbf{W}_{\mathrm{down}},
\end{equation}
any invertible diagonal matrix $\symbox{ffnsym}{\mathbf{S}}$, the reparameterization
\begin{equation}
\mathbf{W}_{\mathrm{up}}'
=
\mathbf{W}_{\mathrm{up}}\symbox{ffnsym}{\mathbf{S}},
\qquad
\mathbf{W}_{\mathrm{down}}'
=
\symbox{ffnsym}{\mathbf{S}}^{-1}\mathbf{W}_{\mathrm{down}}
\end{equation}
satisfies
\begin{equation}
\mathrm{GLU}'(\mathbf{X})=\mathrm{GLU}(\mathbf{X}).
\end{equation}

\section{Matching Algorithms}

Given the functional preserving symmetries introduced in section~\ref{sec:pre}, we now study how to obtain proper symmetries to align two independently trained models. 
We consider a design space of matching approaches, including weight matching and learned matching, and discuss how to parameterize these symmetries to make them learnable through gradient descent.
Building on this, we introduce \textbf{dual learned matching}, which assigns learnable transformations to both checkpoints, yielding a more favorable optimization landscape for finding low-barrier linear paths.

\subsection{From Weight Matching to Learned Matching}

\noindent{\bf Weight Matching.}
Introduced by~\citet{ainsworth2022git}, weight matching formulates the weight alignment problem as maximizing weight similarity under the symmetry family.
The original paper only discusses permutation symmetries in MLP networks, but the same procedure can be extended to more complex symmetries in Transformers.
Following the adoption by~\citet{theus2025generalized}, we first recover the global residual-space symmetry:
\begin{equation}
\symbox{globalsym}{\mathbf{G}}
=
\arg\min_{\mathbf{G}^\top \mathbf{G} = \mathbf{I}}
\left\| \mathbf{R}^A - \mathbf{R}^B \mathbf{G} \right\|_F^2,
\qquad
\symbox{globalsym}{g}
=
\frac{\langle \mathbf{R}^B \mathbf{G}, \mathbf{R}^A \rangle}
{\| \mathbf{R}^B \mathbf{G} \|_F^2}.
\end{equation}
Here $\mathbf{R}^A$ and $\mathbf{R}^B$ are the weights along the residual path collected from both models; they may include the token embedding and all residual read/write matrices. The closed-form solution is given by the SVD of $\mathbf{R}^{B\top}\mathbf{R}^A$.
Then, for each layer $\ell$, we recover the head permutation by comparing the
induced QK and OV circuits. For each head $i$ from model $A$ and head $j$ from model $B$ we define
\begin{equation}
\mathbf{C}^{\mathrm{head}}_{ij}
=
\left\|
\mathbf{W}_{Q,i}^{A}\mathbf{W}_{K,i}^{A\top}
-\mathbf{G}^{\top}\mathbf{W}_{Q,j}^{B}\mathbf{W}_{K,j}^{B\top}\mathbf{G}
\right\|_{F}^{2}
+
\left\|
\mathbf{W}_{V,i}^{A}\mathbf{W}_{O,i}^{A}
-\mathbf{G}^{\top}\mathbf{W}_{V,j}^{B}\mathbf{W}_{O,j}^{B}\mathbf{G}
\right\|_{F}^{2},
\end{equation}
\begin{equation}
\symbox{headsym}{\mathbf{P}_{h,\ell}}
=
\arg\min_{\mathbf{P}\in\Pi_{h}}
\sum_{i,j} P_{ij}\,\mathbf{C}^{\mathrm{head}}_{ij}.
\end{equation}
This compares heads through their functional QK and OV circuits rather than their raw weights, and therefore avoids the ambiguity from per-head internal invertible symmetries. 
After the head permutation is fixed, the remaining QK and OV symmetries inside each head can be solved independently in closed form, so we omit their explicit formulas here.

For FFN layers, we solve the corresponding bilinear assignment problem:
\begin{equation}
\symbox{ffnsym}{\mathbf{P}_{f,\ell}}
=
\arg\max_{\mathbf{P}\in\Pi_{f}}
\left\langle
\mathbf{W}_{1,\ell}^{A},
\mathbf{G}^{\top}\mathbf{W}_{1,\ell}^{B}\mathbf{P}
\right\rangle_{F}
+
\left\langle
\mathbf{W}_{2,\ell}^{A},
\mathbf{P}^{\top}\mathbf{W}_{2,\ell}^{B}\mathbf{G}
\right\rangle_{F}.
\end{equation}
The same construction extends directly to gated FFNs. Since $\mathbf{G}$, head permutations, and FFN permutations are coupled, these matching steps can be iterated, as in the SOBLAP-style procedure by~\citet{ainsworth2022git}. 
After weight matching, the two models are in a substantially better common coordinate system, but weight matching is loss-agnostic and can therefore still leave a visible loss barrier along the linear path.

\noindent{\bf Learned Matching.}
Following~\citet{theus2025generalized}, learned matching refines the weight matching solution by directly optimizing the interpolation loss. Starting from the matched model $\pi(\Theta_B)$, we can do linear interpolation with the other model $\Theta_A$:
\begin{equation}
\Theta_{\mathrm{interp}}
=
\lambda\,\Theta_A
+
(1-\lambda)\,\pi(\Theta_B),
\qquad
\lambda\in[0,1].
\end{equation}
We then optimize the symmetry parameters in $\pi$ to minimize the interpolation loss through gradient descent. 
In this way, learned matching uses task loss to refine the loss-agnostic alignment from weight matching into a lower-barrier linear path.

\subsection{Parameterizing Symmetries}

To optimize continuous symmetry transformations using gradient-based methods, we consider a range of parameterizations for orthogonal and invertible symmetries, differing in how they handle structural constraints during training.

\begin{wraptable}{r}{0.64\textwidth}
\vspace{-15pt}
\caption{Continuous symmetry parameterizations.}
\label{tab:parameterizing-symmetries}
\scriptsize
\setlength{\tabcolsep}{1pt}
\renewcommand{\arraystretch}{0.84}
\begin{tabular}{@{}>{\raggedright\arraybackslash}p{0.18\linewidth}>{\raggedright\arraybackslash}p{0.50\linewidth}>{\raggedright\arraybackslash}p{0.26\linewidth}@{}}
\toprule
Method & Form & Covers \\
\midrule
\textbf{Cayley}
& $\mathbf{Q}=(\mathbf{I}+\mathbf{S})(\mathbf{I}-\mathbf{S})^{-1},\ \mathbf{S}^{\top}=-\mathbf{S}$
& $SO(N)\backslash\{-1\}$ \\
\textbf{Matrix exp.}
& $\mathbf{Q}=\exp(\mathbf{S}),\ \mathbf{S}^{\top}=-\mathbf{S}$
& $SO(N)$ \\
\textbf{Matrix exp.}
& $\mathbf{M}=\exp(\mathbf{U})$
& Subset of $\mathrm{GL}^{+}(N)$  \\
\textbf{Polar}
& $\mathbf{M}=\mathbf{Q}(\mathbf{L}\mathbf{L}^{\top}),\ \mathbf{Q}=\mathrm{Cayley}(\mathbf{S}),$
\newline $\mathbf{L}$ lower-tri., diag$>0$
& Cayley-limited subset of $\mathrm{GL}^{+}(N)$ \\

\bottomrule
\end{tabular}
\vspace{-4pt}
\end{wraptable}

\noindent{\bf Orthogonal.}
For orthogonal symmetries, we primarily consider the Cayley transform and the matrix exponential. The Cayley transform is computationally efficient, numerically stable, and straightforward to implement, but it cannot represent orthogonal matrices whose spectrum contains the eigenvalue $-1$, while the matrix exponential provides a standard parameterization of $SO(N)$ with a higher computational cost. We avoid Householder products due to their relative inefficiency on GPUs, as well as unconstrained parameterizations followed by projection, which exhibited instability during optimization in our experiments.

\noindent{\bf Invertible.}
For invertible symmetries, we explore matrix exponential and polar decomposition style parameterization. The former guarantees invertibility and a positive determinant by setting $\mathbf{M}=\exp(\mathbf{U})$, although it only covers the subset of $\mathrm{GL}^{+}(N)$ that admits a real matrix logarithm. The latter decomposes the transform into a Cayley-parameterized orthogonal factor and a Cholesky-parameterized symmetric positive-definite factor. Additionally, we evaluate unconstrained direct parameterization as a baseline for invertible transforms.

We do not learn discrete symmetries such as FFNs and attention-head permutations, unlike~\citet{theus2025generalized}, as learning them is inefficient for large models and hard to optimize in our setting. These symmetries are kept fixed as the weight matching solution.

\subsection{Dual Learned Matching}
\label{sec:dual_matching}
In both weight matching and learned matching, only one of the two checkpoints is transformed to align with the other, a design choice motivated by simplicity and efficiency. However, this one-sided procedure can be suboptimal, as it requires aligning to a fixed target in the original coordinate system, which may lead to a less favorable optimization landscape. A more natural parameterization is to assign learnable transformations to both checkpoints and optimize them jointly, allowing the models to meet in a shared coordinate system.

We call this approach \textbf{Dual Learned Matching}, and it can be formulated as:
\begin{equation}
\Theta_{\mathrm{interp}}
=
\lambda\,\pi_A(\Theta_A)
+
(1-\lambda)\,\pi_B(\Theta_B),
\qquad
\lambda\in[0,1].
\end{equation}
\begin{wrapfigure}[13]{r}{0.58\textwidth}
\vspace{-8pt}
\small
\refstepcounter{algorithm}
\label{alg:dual_learned_matching}
\addcontentsline{loa}{algorithm}{\protect\numberline{\thealgorithm}{Dual learned matching (LMC-DM)}}
\hrule height .8pt depth 0pt \kern 2pt
\noindent\textbf{Algorithm~\thealgorithm} Dual learned matching (LMC-DM)
\kern 2pt \hrule \kern 2pt
\begin{algorithmic}[1]
\Require Checkpoints $\Theta_A,\Theta_B$; dataset $\mathcal{D}$; iterations $T$; learning rate $\eta$.
\State Initialize $\Pi_A,\Pi_B$ from weight matching.
\Statex $\Pi_m=(g_m,\mathbf{G}_m,\mathbf{M}_{QK,m},\mathbf{M}_{OV,m}),\quad m\in\{A,B\}$.
\Statex $\mathbf{G}_m,\mathbf{M}_{QK,m},\mathbf{M}_{OV,m}\in\{\textsc{Cayley},\textsc{Polar},\textsc{Exp}\}$.
\For{$t=1$ \textbf{to} $T$}
    \State $\Theta_A' \leftarrow \pi_{\Pi_A}(\Theta_A)$, \quad $\Theta_B' \leftarrow \pi_{\Pi_B}(\Theta_B)$.
    \State Sample $\lambda \sim \mathcal{U}(0,1)$ and a minibatch $B \sim \mathcal{D}$.
    \State Interpolate: $\Theta_{\mathrm{interp}} \leftarrow \lambda\,\Theta_A' + (1-\lambda)\,\Theta_B'$.
    \State Compute $\mathcal{J} \leftarrow \frac{1}{|B|}\sum_{(x,y)\in B}\mathcal{L}(\Theta_{\mathrm{interp}};x,y)$.
    \State Update $\Pi_A,\Pi_B$ by gradient descent on $\mathcal{J}$.
\EndFor
\State \Return $\Pi_A,\Pi_B$.
\end{algorithmic}
\kern 2pt \hrule
\vspace{-2pt}
\end{wrapfigure}
\noindent
Starting from the weight matching solution, we optimize the transformations on both checkpoints directly through interpolation loss. In practice, we keep the discrete head and FFN permutations fixed and optimize the continuous symmetries on both sides, so that both models can move toward a shared coordinate system rather than forcing one model to match the other exactly. Allowing these discrete permutations to learn does not flip any assignment relative to the weight-matching solution or change the barrier, so we keep them fixed.
\WFclear

\section{Experiments}
\label{sec:exp}

\subsection{Experimental Setup}
We use publicly available independently trained checkpoints with identical architectures, including Vision Transformers~\cite{dosovitskiy2020image} and language models~\cite{biderman2023pythia,Groeneveld2023OLMo}, to evaluate our method. 
We assess linear mode connectivity by sampling points along the linear interpolation between two endpoints and measuring the loss along the path. 
We report the loss barrier~\cite{frankle2020linear}:
\begin{equation}
\max_{\lambda}\mathcal{L}(\Theta_\lambda)
-
\frac{1}{2}\bigl(\mathcal{L}(\Theta_A)+\mathcal{L}(\Theta_B)\bigr)
\end{equation}
approximated over the sampled path.
For ViTs, we also evaluate ImageNet-1K~\cite{deng2009imagenet} top-1 accuracy along the interpolation path, using interpolation intervals of 0.1. 
For language models, we report WikiText~\cite{merity2016pointer} perplexity additionally. 
Weight Matching (WM) and Learned Matching (LM) are methods from~\citet{ainsworth2022git} and~\citet{theus2025generalized}, respectively. 
LMC-DM denotes our dual learned matching procedure with learnable parameterization of continuous symmetries. 
Enhanced learned matching refers to the same procedure without dual parameterization. 
We use Cayley parameterization for orthogonal symmetries and polar-style parameterization for invertible symmetries by default for our methods. We consider two initialization schemes based on weight matching (WM): an absorbed initialization, where the WM solution is absorbed into the endpoints and the symmetries are initialized as identity, and a non-absorbed initialization, where the symmetries are initialized with the WM solution. The two schemes are equivalent at initialization in terms of the resulting interpolation.
Additional experimental details are provided in Appendix~\ref{app:exp-details}.

\subsection{Main Results}

\begin{table}[t]
\centering
\caption{Loss barrier comparison across ViT and Pythia models. Lower is better.}
\label{tab:main-results}
\small
\setlength{\tabcolsep}{3pt}
\renewcommand{\arraystretch}{1.08}
\begin{tabularx}{\textwidth}{l *{7}{>{\centering\arraybackslash}X}}
\toprule
\multirow{2}{*}{Method} &
\multicolumn{3}{c}{ViT (ImageNet-1K)} &
\multicolumn{4}{c}{Pythia (WikiText)} \\
\cmidrule(lr){2-4}\cmidrule(lr){5-8}
& Small & Base & Large & 14M & 160M & 410M & 1B \\
\midrule
Raw Interpolation & 6.58 & 6.57 & 6.47 & 9.89 & 6.28 & 4.99 & 4.60 \\
Weight Matching~\cite{ainsworth2022git} & 6.58 & 6.50 & 6.48 & 8.17 & 6.04 & 5.65 & 4.75 \\
Learned Matching~\cite{theus2025generalized} & 2.18 & 2.00 & 1.23 & 1.26 & 0.39 & 0.57 & 0.42 \\
Enhanced Learned Matching (Ours) & 1.20 & 1.05 & 0.71 & 1.00 & 0.27 & 0.44 & 0.28 \\
\rowcolor{black!6}
\textbf{Dual Learned Matching (Ours)} & \textbf{1.11} & \textbf{0.82} & \textbf{0.66} & \textbf{0.86} & \textbf{0.18} & \textbf{0.37} & \textbf{0.23} \\
\bottomrule
\end{tabularx}
\vspace{-0.15in}
\end{table}

Table~\ref{tab:main-results} summarizes the loss-barrier comparison across ViT and Pythia scales. 
Simple WM performs poorly under modern vision and language models.
While LM substantially reduces the barrier compared to WM, it still leaves a significant gap to zero-barrier connectivity. Our enhanced version of learned matching further reduces the barrier.
LMC-DM achieves the lowest barrier among all methods, indicating that jointly optimizing both endpoints yields better interpolation paths than one-sided learned matching.

\begin{figure}[t]
\centering
\setlength{\abovecaptionskip}{3pt}
\setlength{\belowcaptionskip}{-3pt}
\includegraphics[width=\linewidth]{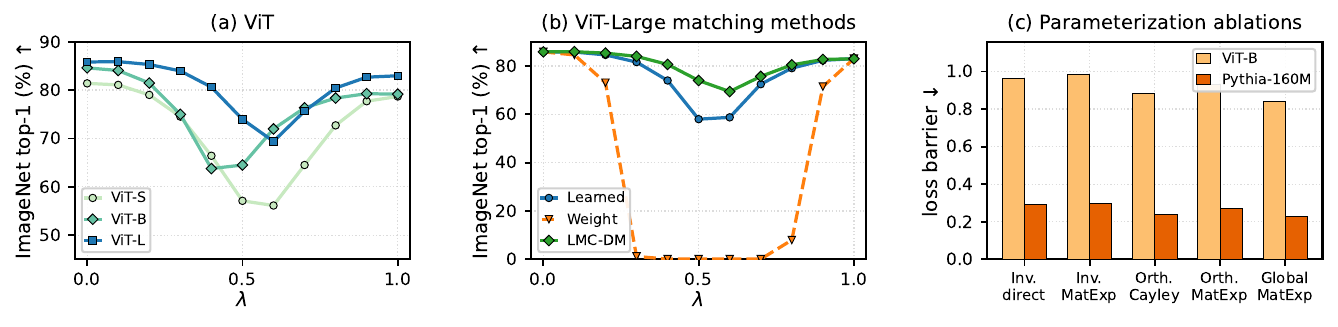}
\vspace{-1.9em}
\caption{Summary of interpolation and parameterization results. (a) LMC-DM across ViT-Small, ViT-Base, and ViT-Large. (b) ViT-Large accuracy paths under learned matching, weight matching, and LMC-DM. (c) Loss-barrier comparison across symmetry parameterizations on ViT-Base and Pythia-160M.}
\label{fig:exp-vit-param-panels}
\vspace{-0.1in}
\end{figure}

\subsection{Ablation Studies}

\noindent{\bf Symmetry Components.}
We first ablate the symmetry components used in LMC-DM, as shown in Table~\ref{tab:ablation-side-by-side}. 
Disabling the QK or OV symmetries leads to a significantly higher barrier, especially for ViTs. 
For Pythia, disabling QK symmetries has a smaller impact. 
Disabling both symmetries further increases the barrier, but still performs better than standard learned matching, which does not leverage attention symmetries and uses a single-sided parameterization. 
The global scaling symmetry has a smaller but non-negligible effect.

\noindent{\bf Symmetry Parameterization.}
We then compare different parameterizations for continuous symmetries. 
Despite having fewer degrees of freedom, Cayley-based orthogonal parameterization outperforms unconstrained and matrix-exponential parameterizations for invertible symmetries. 
This may be because the Cayley transform is more numerically stable and easier to optimize than the matrix exponential. 
Combined with Cayley, the polar-style  parameterization for invertible symmetries achieves the best performance, balancing both stability and expressivity.

\begin{wraptable}{r}{0.48\textwidth}
\vspace{-17pt}
\centering
\caption{Single vs. dual endpoint matching.}
\label{tab:ablation-barriers}
\footnotesize
\setlength{\tabcolsep}{3pt}
\begin{tabular*}{0.43\textwidth}{@{\extracolsep{\fill}}lcc}
\toprule
Setting & ViT-B $\downarrow$ & Pythia-160M $\downarrow$ \\
\midrule
Single \textit{w. Absorb} & 1.05 & 0.27 \\
Single \textit{w/o Absorb} & 1.47 & 0.41 \\
Dual \textit{w/o Absorb} & 0.82 & 0.21 \\
Dual \textit{w. Absorb} & \textbf{0.82} & \textbf{0.18} \\
\bottomrule
\end{tabular*}
\vspace{-8pt}
\end{wraptable}
\noindent{\bf Single vs. Dual Matching.}
We compare whether the matching transformation should be learned on one endpoint or on both endpoints. Dual matching gives both checkpoints room to move toward a shared basin, which consistently lowers the barrier compared
with the single-sided variants. Absorbing the weight-matching solution
into the endpoints leverages clean identical initialization for symmetries, which further reduces the barrier.

\begin{table}[!t]
\centering
\begin{minipage}[t]{0.56\textwidth}
\vspace{0pt}
\centering
\includegraphics[width=\linewidth]{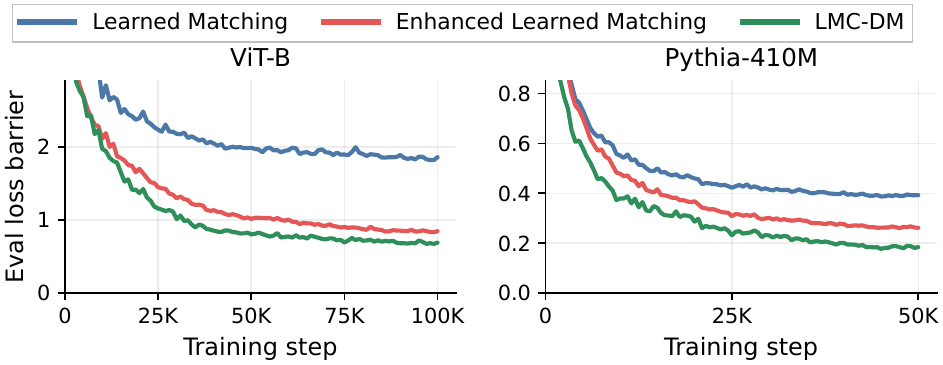}
\end{minipage}\hfill
\begin{minipage}[t]{0.40\textwidth}
\vspace{0pt}
\refstepcounter{figure}\label{fig:vitb-pythia410m-eval-barriers}
\small
\textbf{Figure~\thefigure: Evaluation Loss Barriers.}
We track ViT-B and Pythia-410M during matching. LMC-DM reduces barriers faster and reaches lower final barriers than learned matching and the enhanced learned matching. For efficiency, these evaluation barriers are measured on a subset rather than the full test or validation set.
\end{minipage}

\vspace{7pt}
\caption{Ablation study on ViT-Base and Pythia-160M. Small lower-right numbers report the raw numerical change from the full LMC-DM baseline; green indicates better and red indicates worse.}
\label{tab:ablation-side-by-side}
\small
\setlength{\tabcolsep}{2.4pt}
\newcommand{\abdelta}[2]{\raisebox{-0.55ex}{\textcolor{#1}{\scriptsize\,#2}}}
\newcommand{\abcell}[3]{\mbox{#1\abdelta{#2}{#3}}}
\begin{tabular*}{\textwidth}{@{\extracolsep{\fill}}>{\raggedright\arraybackslash}p{0.30\textwidth}llllll}
\toprule
\multirow{2}{*}{Ablation} & \multicolumn{3}{c}{ViT-Base} & \multicolumn{3}{c}{Pythia-160M} \\
\cmidrule(lr){2-4}\cmidrule(lr){5-7}
& Barrier $\downarrow$ & Worst Acc $\uparrow$ & Mid. Acc $\uparrow$ & Barrier $\downarrow$ & Worst PPL $\downarrow$ & Mid. PPL $\downarrow$ \\
\midrule
Ours \textit{(Dual, Global Cayley,}\\[-1pt]\quad \textit{Attention Invertible Polar)} &
0.82 & 63.79 & 65.48 & 0.18 & 82.47 & 73.33 \\
\addlinespace[1pt]
\multicolumn{7}{@{}l}{\textit{Component removals}} \\
\quad w/o. QK transform & \abcell{0.97}{red!70!black}{+0.15} & \abcell{60.68}{red!70!black}{-3.11} & \abcell{61.80}{red!70!black}{-3.68} & \abcell{0.18}{black!55}{+0.00} & \abcell{81.81}{green!45!black}{-0.66} & \abcell{76.85}{red!70!black}{+3.52} \\
\quad w/o. OV transform & \abcell{1.17}{red!70!black}{+0.35} & \abcell{56.85}{red!70!black}{-6.94} & \abcell{61.04}{red!70!black}{-4.44} & \abcell{0.37}{red!70!black}{+0.19} & \abcell{97.03}{red!70!black}{+14.56} & \abcell{82.97}{red!70!black}{+9.64} \\
\quad w/o. QK and OV & \abcell{1.48}{red!70!black}{+0.66} & \abcell{50.98}{red!70!black}{-12.81} & \abcell{54.96}{red!70!black}{-10.52} & \abcell{0.37}{red!70!black}{+0.19} & \abcell{97.06}{red!70!black}{+14.59} & \abcell{82.11}{red!70!black}{+8.78} \\
\quad w/o. global scale & \abcell{0.82}{black!55}{+0.00} & \abcell{63.50}{red!70!black}{-0.29} & \abcell{65.39}{red!70!black}{-0.09} & \abcell{0.23}{red!70!black}{+0.05} & \abcell{86.01}{red!70!black}{+3.54} & \abcell{77.53}{red!70!black}{+4.20} \\
\addlinespace[1pt]
\multicolumn{7}{@{}l}{\textit{Attention parameterization}} \\
\quad Invertible, direct & \abcell{0.97}{red!70!black}{+0.15} & \abcell{60.72}{red!70!black}{-3.07} & \abcell{63.68}{red!70!black}{-1.80} & \abcell{0.29}{red!70!black}{+0.11} & \abcell{90.20}{red!70!black}{+7.73} & \abcell{80.31}{red!70!black}{+6.98} \\
\quad Invertible, MatExp & \abcell{0.98}{red!70!black}{+0.16} & \abcell{60.26}{red!70!black}{-3.53} & \abcell{63.57}{red!70!black}{-1.91} & \abcell{0.30}{red!70!black}{+0.12} & \abcell{90.59}{red!70!black}{+8.12} & \abcell{78.06}{red!70!black}{+4.73} \\
\quad Orthogonal, Cayley & \abcell{0.88}{red!70!black}{+0.06} & \abcell{62.42}{red!70!black}{-1.37} & \abcell{63.62}{red!70!black}{-1.86} & \abcell{0.24}{red!70!black}{+0.06} & \abcell{85.18}{red!70!black}{+2.71} & \abcell{71.86}{green!45!black}{-1.47} \\
\quad Orthogonal, MatExp & \abcell{0.93}{red!70!black}{+0.11} & \abcell{61.53}{red!70!black}{-2.26} & \abcell{63.26}{red!70!black}{-2.22} & \abcell{0.27}{red!70!black}{+0.09} & \abcell{88.10}{red!70!black}{+5.63} & \abcell{73.99}{red!70!black}{+0.66} \\
\addlinespace[1pt]
\multicolumn{7}{@{}l}{\textit{Global parameterization}} \\
\quad MatExp & \abcell{0.84}{red!70!black}{+0.02} & \abcell{63.33}{red!70!black}{-0.46} & \abcell{65.63}{green!45!black}{+0.15} & \abcell{0.23}{red!70!black}{+0.05} & \abcell{86.00}{red!70!black}{+3.53} & \abcell{76.45}{red!70!black}{+3.12} \\
\bottomrule
\end{tabular*}
\vspace{-0.15in}
\end{table}

\subsection{Scaling Up to Larger Models}
We apply our method to Pythia and OLMo models up to the billion-parameter scale. 
As model size increases, the loss barrier for Pythia remains low, even at 6.9B, while OLMo-7B exhibits a significantly higher barrier. 
These results indicate that large pretrained Transformers can still exhibit linear mode connectivity. 
At the same time, the discrepancy between Pythia and OLMo suggests that LMC in large models is more nuanced and influenced by factors beyond scale, such as training data, initialization, and optimization dynamics.

\section{Related Work}

\noindent{\bf Linear Mode Connectivity.}
Early work~\cite{garipov2018loss,draxler2018essentially} introduced mode connectivity by showing that independently trained networks can be connected through stable low-loss paths. Follow-up studies~\cite{tatro2020optimizing,entezari2021role} showed that high linear interpolation barriers often arise from unresolved permutation symmetries, and that resolving them can largely recover linear mode connectivity (LMC). \citet{frankle2020linear} further connected LMC to the lottery ticket hypothesis, showing that lottery tickets can lie in a shared linearly connected basin. More recent work extends LMC to layer-wise connectivity, broader symmetry groups, multi-model settings, and modern architectures such as Transformers and MoEs~\cite{zhou2023going,adilova2023layer,theus2025generalized,ito2025linear,tran2025linear}.

\noindent{\bf Weight Symmetries and Invariances.} This refers to the family of transformations of model parameters that exactly preserve the functionality. 
Neuron permutation symmetries in MLPs have long been known~\cite{hecht1990algebraic,chen1993geometry}. 
\citet{ashkboos2024slicegpt} observed orthogonal symmetries in the residual space of Transformers, along with absorbable symmetries between normalization and linear layers. 
\citet{zhang2025beyond,theus2025generalized} identified head-wise symmetries in Transformers. 
A systematic survey of parameter symmetries is provided by \citet{zhao2025symmetry}.

\noindent{\bf Pretrained Model Merging.}
Pretrained model merging aims to combine multiple independently trained checkpoints into a single model, and is significantly more challenging than task arithmetic in the fine-tuning regime, as independently trained models often diverge into different basins~\cite{frankle2020linear,ilharco2022editing,yang2026model}. 
OT-Fusion~\cite{singh2020model} first leveraged optimal transport to align neurons, enabling data-driven model merging that outperforms naive interpolation. 
Subsequent work, such as Git-Rebasin~\cite{ainsworth2022git}, proposed practical approaches to align models under permutation symmetries to achieve low-barrier fusion. 
More recent studies have extended these ideas to pretrained Transformers~\cite{verma2024merging,zhang2025beyond,theus2025generalized}.
\vspace{-0.05in}
\section{Conclusion}
\vspace{-0.05in}

We studied whether and to what extent linear mode connectivity can be achieved in modern pretrained Transformers by accounting for their weight-space symmetries. 
We explicitly parameterize the relevant symmetries in modern Transformers and optimize them with dual learned matching, which allows both endpoints to be symmetrically aligned toward a shared linear interpolation path.
This alignment places independently trained checkpoints in a shared coordinate system, enabling meaningful linear interpolation across ViTs and LLMs up to the billion-parameter scale. 
More broadly, this perspective suggests that modern large models are not isolated endpoints in weight space, but rather symmetry-related representations that can be aligned and connected. Future work includes extending this framework to broader architectures and training settings, applying it to practical model merging and ensembling, and further interpreting the geometry of the loss landscape under these symmetries. 
Limitations and further discussion can be found in Appendix~\ref{app:limitations} and Appendix~\ref{app:further_discussion}.



\bibliography{main}
\bibliographystyle{unsrtnat}

\clearpage
\appendix



\appendix

\section*{\Large{Appendix}}

\section{Limitations}
\label{app:limitations}

Our framework relies on explicitly parameterized functionality-preserving symmetries, and therefore may not capture all sources of misalignment between independently pretrained models. The current formulation mainly focuses on models with compatible architectures, dimensions, and tokenization spaces, extending it to heterogeneous architectures or models trained with substantially different objectives remains challenging.
Besides, in some Transformer architectures, the available functionality-preserving symmetries are structurally limited. 
In such cases, the effectiveness of our method may be limited due to the reduced symmetry capacity. 

\section{Societal Impacts}

This work may have positive societal impacts by making large pretrained models easier to reuse, combine, and analyze. If independently trained Transformers can be aligned and merged through low-loss linear paths, practitioners may be able to consolidate useful capabilities from multiple checkpoints without full retraining, reducing computational cost, energy consumption, and barriers to model development. The proposed symmetry-based analysis may also improve transparency by providing a more structured understanding of how different models relate in weight space, which could support safer model editing, auditing, and deployment.

The work may also have negative societal impacts. More effective model merging could make it easier to combine capabilities from different models, including potentially harmful capabilities, without extensive training resources. This may lower the barrier for creating models with stronger misuse potential or for obscuring the provenance of merged systems. In addition, if merged models are deployed without sufficient evaluation, they may inherit or amplify biases, unsafe behaviors, or failure modes from their source models. 
Moreover, if model merging becomes easier or more widely accessible, it may also increase the risk of intellectual property leakage. Proprietary capabilities, memorized content, or model-specific behaviors from a source checkpoint may be transferred to a merged model, while making attribution and ownership more difficult to verify. As a result, model owners may have reduced incentives to openly release their checkpoints due to concerns about uncontrolled reuse and merging.

\section{Discussion: Why Does Dual Learned Matching Work?}
\label{app:single_dual_equiv}

A notable and perhaps surprising fact is that any dual parameterized symmetry can be equivalently absorbed into a single-sided parameterization. We provide a formal proof of this equivalence in Appendix~\ref{app:absorb_proof}. This result implies that dual parameterization does not introduce additional degrees of freedom, despite its empirical advantage over single-sided optimization. Therefore, the improved performance of dual learned matching cannot be attributed to increased parameter or symmetry capacity, but rather to the more favorable optimization geometry induced by overparameterization.

One hypothesis is that dual parameterization enables both checkpoints to move toward a shared, smoother region of the loss landscape, whereas single-sided optimization restricts movement to only one checkpoint, effectively anchoring the other. Moreover, dual parameterization allows complex symmetry transformations between two models to be decomposed into two simpler and more tractable transformations, thereby facilitating optimization.

\subsection{Proof of Dual-to-Single Absorption}
\label{app:absorb_proof}
We prove the equivalence by showing that the two endpoint transformations can be rewritten as one relative transformation plus a common functionality-preserving gauge. We first illustrate the reduction using the global residual-space symmetry. Let $\mathbf{G}_A,\mathbf{G}_B$ be the global orthogonal transformations applied to endpoints $A$ and $B$. Under the row-vector convention, a residual-stream representation transforms as $\mathbf{X} \mapsto \mathbf{X}\mathbf{G}$, so residual-reading and residual-writing matrices transform as
\[
\mathbf{W}_{\mathrm{in}} \mapsto \mathbf{G}^\top \mathbf{W}_{\mathrm{in}},
\qquad
\mathbf{W}_{\mathrm{out}} \mapsto \mathbf{W}_{\mathrm{out}}\mathbf{G} .
\]
The dual interpolation gives
\[
\widetilde{\mathbf{W}}_{\mathrm{in}}
=
\lambda \mathbf{G}_A^\top \mathbf{W}_{\mathrm{in}}^A
+
(1-\lambda)\mathbf{G}_B^\top \mathbf{W}_{\mathrm{in}}^B,
\qquad
\widetilde{\mathbf{W}}_{\mathrm{out}}
=
\lambda \mathbf{W}_{\mathrm{out}}^A \mathbf{G}_A
+
(1-\lambda)\mathbf{W}_{\mathrm{out}}^B \mathbf{G}_B .
\]
Define the relative global transformation
\[
\mathbf{R}_{G} := \mathbf{G}_B \mathbf{G}_A^\top .
\]
Then $\mathbf{G}_B=\mathbf{R}_{G}\mathbf{G}_A$ and $\mathbf{G}_B^\top=\mathbf{G}_A^\top \mathbf{R}_{G}^\top$, so
\[
\widetilde{\mathbf{W}}_{\mathrm{in}}
=
\mathbf{G}_A^\top
\underbrace{
\big(
\lambda \mathbf{W}_{\mathrm{in}}^A
+
(1-\lambda)\mathbf{R}_{G}^\top \mathbf{W}_{\mathrm{in}}^B
\big)
}_{\text{single-sided relative interpolation}},
\]
\[
\widetilde{\mathbf{W}}_{\mathrm{out}}
=
\underbrace{
\big(
\lambda \mathbf{W}_{\mathrm{out}}^A
+
(1-\lambda)\mathbf{W}_{\mathrm{out}}^B \mathbf{R}_{G}
\big)
}_{\text{single-sided relative interpolation}}
\mathbf{G}_A .
\]
Because $\mathbf{G}_A$ and $\mathbf{G}_B$ are both orthogonal, their product $\mathbf{R}_{G}=\mathbf{G}_B\mathbf{G}_A^\top$ is also orthogonal. Thus the relative transformation $\mathbf{R}_{G}$ has exactly the same parameterized form as the original global residual-space symmetry parameter $\mathbf{G}$. Therefore the dual-interpolated layer is exactly the single-sided relative-interpolated layer under a valid global symmetry $\mathbf{R}_{G}$, followed by the same common global gauge $\mathbf{G}_A$ on the residual stream. Since this gauge is applied consistently to all residual reads and writes, it preserves the model function. Thus dual global matching is equivalent to single-sided relative global matching along the entire interpolation path.

For local circuit symmetries, the cancellation can be shown within each attention head by multiplying the paired matrices. For example, for the OV circuit of head $i$, define
\[
\mathbf{R}_{V,i} := \mathbf{M}_{V,i}^{B}(\mathbf{M}_{V,i}^{A})^{-1}.
\]
Then the dual interpolation can be rewritten as
\[
\widetilde{\mathbf{W}}_{V,i}
=
\big(
\lambda \mathbf{W}_{V,i}^A
+
(1-\lambda)\mathbf{W}_{V,i}^B \mathbf{R}_{V,i}
\big)\mathbf{M}_{V,i}^{A},
\qquad
\widetilde{\mathbf{W}}_{O,i}
=
(\mathbf{M}_{V,i}^{A})^{-1}
\big(
\lambda \mathbf{W}_{O,i}^A
+
(1-\lambda)\mathbf{R}_{V,i}^{-1}\mathbf{W}_{O,i}^B
\big),
\]
and hence
\[
\widetilde{\mathbf{W}}_{V,i}\widetilde{\mathbf{W}}_{O,i}
=
\big(
\lambda \mathbf{W}_{V,i}^A
+
(1-\lambda)\mathbf{W}_{V,i}^B \mathbf{R}_{V,i}
\big)
\underbrace{\mathbf{M}_{V,i}^{A}(\mathbf{M}_{V,i}^{A})^{-1}}_{\mathbf{I}}
\big(
\lambda \mathbf{W}_{O,i}^A
+
(1-\lambda)\mathbf{R}_{V,i}^{-1}\mathbf{W}_{O,i}^B
\big).
\]
The same relative-parameter argument applies to the QK circuit in each head with
\[
\mathbf{R}_{Q,i} := \mathbf{M}_{Q,i}^{B}(\mathbf{M}_{Q,i}^{A})^{-1},
\]
and to head and FFN permutations with
\[
\mathbf{P}_{h,\ell} := \mathbf{P}_{h,\ell}^{B}(\mathbf{P}_{h,\ell}^{A})^\top,
\qquad
\mathbf{P}_{f,\ell} := \mathbf{P}_{f,\ell}^{B}(\mathbf{P}_{f,\ell}^{A})^\top .
\]
Thus the dual-sided symmetry parameterization used in our method can be absorbed into a single-sided relative parameterization.

\section{Experimental Details}
\label{app:exp-details}
\begin{table}[h]
\centering
\caption{Endpoint checkpoints and architecture used in our experiments. Each endpoint pair in a row shares the listed architecture; $d_h$ is the attention head dimension.}
\label{tab:model-metadata}
\scriptsize
\setlength{\tabcolsep}{0.8pt}
\renewcommand{\arraystretch}{1.08}
\begin{tabularx}{\textwidth}{@{}>{\raggedright\arraybackslash}p{0.11\textwidth}>{\raggedright\arraybackslash}X>{\centering\arraybackslash}p{0.08\textwidth}>{\centering\arraybackslash}p{0.04\textwidth}>{\centering\arraybackslash}p{0.052\textwidth}>{\centering\arraybackslash}p{0.06\textwidth}>{\centering\arraybackslash}p{0.045\textwidth}>{\centering\arraybackslash}p{0.095\textwidth}>{\centering\arraybackslash}p{0.052\textwidth}@{}}
\toprule
Model & Hugging Face Checkpoint & Arch. & $L$ & $d$ & $d_{\mathrm{ff}}$ & $d_h$ & Norm. & Bias \\
\midrule
\multirow{2}{*}{ViT-S} &
A: \path{timm/vit_small_patch16_224.augreg_in21k_ft_in1k} &
\multirow{2}{*}{ViT} & \multirow{2}{*}{12} & \multirow{2}{*}{384} & \multirow{2}{*}{1536} & \multirow{2}{*}{64} & \multirow{2}{*}{LN} & \multirow{2}{*}{Yes} \\
& B: \path{timm/vit_small_patch16_224.augreg_in1k} & & & & & & & \\
\multirow{2}{*}{ViT-B} &
A: \path{timm/vit_base_patch16_224.augreg_in21k_ft_in1k} &
\multirow{2}{*}{ViT} & \multirow{2}{*}{12} & \multirow{2}{*}{768} & \multirow{2}{*}{3072} & \multirow{2}{*}{64} & \multirow{2}{*}{LN} & \multirow{2}{*}{Yes} \\
& B: \path{timm/vit_base_patch16_224.augreg_in1k} & & & & & & & \\
\multirow{2}{*}{ViT-L} &
A: \path{timm/vit_large_patch16_224.augreg_in21k_ft_in1k} &
\multirow{2}{*}{ViT} & \multirow{2}{*}{24} & \multirow{2}{*}{1024} & \multirow{2}{*}{4096} & \multirow{2}{*}{64} & \multirow{2}{*}{LN} & \multirow{2}{*}{Yes} \\
& B: \path{glasses/vit_large_patch16_224} & & & & & & & \\
\midrule
\multirow{2}{*}{Pythia-14M} &
A: \path{EleutherAI/pythia-14m} &
\multirow{2}{*}{GPT-NeoX} & \multirow{2}{*}{6} & \multirow{2}{*}{128} & \multirow{2}{*}{512} & \multirow{2}{*}{32} & \multirow{2}{*}{LN} & \multirow{2}{*}{Yes} \\
& B: \path{EleutherAI/pythia-14m-deduped} & & & & & & & \\
\multirow{2}{*}{Pythia-70M} &
A: \path{EleutherAI/pythia-70m} &
\multirow{2}{*}{GPT-NeoX} & \multirow{2}{*}{6} & \multirow{2}{*}{512} & \multirow{2}{*}{2048} & \multirow{2}{*}{64} & \multirow{2}{*}{LN} & \multirow{2}{*}{Yes} \\
& B: \path{EleutherAI/pythia-70m-deduped} & & & & & & & \\
\multirow{2}{*}{Pythia-160M} &
A: \path{EleutherAI/pythia-160m} &
\multirow{2}{*}{GPT-NeoX} & \multirow{2}{*}{12} & \multirow{2}{*}{768} & \multirow{2}{*}{3072} & \multirow{2}{*}{64} & \multirow{2}{*}{LN} & \multirow{2}{*}{Yes} \\
& B: \path{EleutherAI/pythia-160m-deduped} & & & & & & & \\
\multirow{2}{*}{Pythia-410M} &
A: \path{EleutherAI/pythia-410m} &
\multirow{2}{*}{GPT-NeoX} & \multirow{2}{*}{24} & \multirow{2}{*}{1024} & \multirow{2}{*}{4096} & \multirow{2}{*}{64} & \multirow{2}{*}{LN} & \multirow{2}{*}{Yes} \\
& B: \path{EleutherAI/pythia-410m-deduped} & & & & & & & \\
\multirow{2}{*}{Pythia-1B} &
A: \path{EleutherAI/pythia-1b} &
\multirow{2}{*}{GPT-NeoX} & \multirow{2}{*}{16} & \multirow{2}{*}{2048} & \multirow{2}{*}{8192} & \multirow{2}{*}{256} & \multirow{2}{*}{LN} & \multirow{2}{*}{Yes} \\
& B: \path{EleutherAI/pythia-1b-deduped} & & & & & & & \\
\multirow{2}{*}{Pythia-1.4B} &
A: \path{EleutherAI/pythia-1.4b} &
\multirow{2}{*}{GPT-NeoX} & \multirow{2}{*}{24} & \multirow{2}{*}{2048} & \multirow{2}{*}{8192} & \multirow{2}{*}{128} & \multirow{2}{*}{LN} & \multirow{2}{*}{Yes} \\
& B: \path{EleutherAI/pythia-1.4b-deduped} & & & & & & & \\
\multirow{2}{*}{Pythia-6.9B} &
A: \path{EleutherAI/pythia-6.9b} &
\multirow{2}{*}{GPT-NeoX} & \multirow{2}{*}{32} & \multirow{2}{*}{4096} & \multirow{2}{*}{16384} & \multirow{2}{*}{128} & \multirow{2}{*}{LN} & \multirow{2}{*}{Yes} \\
& B: \path{EleutherAI/pythia-6.9b-deduped} & & & & & & & \\
\midrule
\multirow{2}{*}{HF-1.8B} &
A: \texttt{HuggingFaceFW/}\texttt{ablation-\allowbreak model-\allowbreak fineweb-\allowbreak edu} &
\multirow{2}{*}{LLaMA} & \multirow{2}{*}{24} & \multirow{2}{*}{2048} & \multirow{2}{*}{8192} & \multirow{2}{*}{64} & \multirow{2}{*}{RMSNorm} & \multirow{2}{*}{No} \\
& B: \texttt{HuggingFaceFW/}\texttt{ablation-\allowbreak model-\allowbreak the-\allowbreak pile} & & & & & & & \\
\multirow{2}{*}{OLMo-7B} &
A: \path{allenai/OLMo-7B-hf} &
\multirow{2}{*}{OLMo} & \multirow{2}{*}{32} & \multirow{2}{*}{4096} & \multirow{2}{*}{11008} & \multirow{2}{*}{128} & \multirow{2}{*}{\shortstack{LN\\(non-param.)}} & \multirow{2}{*}{No} \\
& B: \path{allenai/OLMo-7B-Twin-2T-hf} & & & & & & & \\
\bottomrule
\end{tabularx}
\end{table}

We provided the details of the original pretrained models used in our study in Table~\ref{tab:model-metadata}. For each experiment, both models are from open-source checkpoints available on Hugging Face.
For the ViT models, we adopt the \texttt{timm} version of the Google Research AugReg ViT checkpoints.
The ImageNet-21k variants are pretrained on ImageNet-21k and fine-tuned on ImageNet-1k, whereas the ImageNet-1k variants are pretrained solely on ImageNet-1k.
For the Pythia series, each model size is trained on the Pile~\cite{gao2020pile} for approximately 300B tokens by EleutherAI; one checkpoint is trained on the original dataset, while the other is trained on a deduplicated version.
For the HuggingFaceFW 1.8B ablation models, both checkpoints are trained for 350B tokens—one on the FineWeb-Edu~\cite{lozhkov2024fineweb-edu} dataset and the other on the Pile.

For symmetry learning, we optimize the parameters using AdamW with a cosine learning rate scheduler. For ViT models, training is conducted on ImageNet-1K with a batch size of 64. The learning rate starts at 1e-4 and gradually decreases to 2e-5 over 100,000 steps, equivalent to approximately 5 epochs. For language models, training is performed on WikiText-103~\cite{merity2016pointer} with a batch size of 16 and a sequence length of 512. For the 14M-1.4B versions of Pythia, we use an initial learning rate of 5e-5, decreasing to 1e-5 over 50,000 steps. For Pythia-6.9B, the learning rate decreases from 3e-5 to 1e-5 over 20,000 steps, while for OLMo-7B, it decreases from 2e-5 to 5e-6 over 20,000 steps.

All models are trained using bf16 mixed precision, except for Pythia-14M and Pythia-70M, which used fp32 due to the instability of bf16. This instability may stem from unusually large logits, a known issue report in their official repository. For perplexity evaluation, we use a sequence length of 512 without a sliding window and reported the results on the WikiText-103 test set.

The entire symmetry learning procedure is lightweight. ViT and smaller Pythia models are trained on a single A6000 GPU for several hours to one day. For larger models, such as Pythia-6.9B and OLMo-7B, we utilize four A6000 GPUs, with training taking approximately one day.

The interpolation paths across model scales after LMC-DM are reported in Table~\ref{tab:scale-interpolation-paths}. We also report the raw interpolation paths before matching in Table~\ref{tab:raw-interpolation-paths} as unaligned baselines.

\begin{table}[h]
\centering
\caption{Interpolation paths across model scales after dual learned matching. ViT rows report ImageNet-1K accuracy (\%), language-model rows report WikiText perplexity, and the rightmost column reports loss barrier.}
\label{tab:scale-interpolation-paths}
\footnotesize
\setlength{\tabcolsep}{1.8pt}
\renewcommand{\arraystretch}{1.08}
\begin{tabularx}{\textwidth}{@{}>{\raggedright\arraybackslash}p{0.14\textwidth}*{11}{>{\centering\arraybackslash}X}@{\hspace{4pt}}|@{\hspace{4pt}}>{\centering\arraybackslash}X@{}}
\toprule
\multirow{2}{*}{Model} & \multicolumn{11}{c}{Interpolation coefficient $\lambda$} & \multirow{2}{*}{\shortstack{Loss\\Barrier}} \\
\cmidrule(lr){2-12}
& $0.0$ & $0.1$ & $0.2$ & $0.3$ & $0.4$ & $0.5$ & $0.6$ & $0.7$ & $0.8$ & $0.9$ & $1.0$ & \\
\midrule
\rowcolor{black!4}
\multicolumn{13}{@{}l}{\textit{ViT, Accuracy (\%)}} \\
\quad Small & 81.40 & 81.08 & 79.01 & 74.52 & 66.42 & 57.10 & 56.13 & 64.51 & 72.71 & 77.71 & 78.68 & 1.11 \\
\quad Base & 84.62 & 84.05 & 81.46 & 74.99 & 63.77 & 64.51 & 71.95 & 76.32 & 78.39 & 79.25 & 79.14 & 0.82 \\
\quad Large & 85.82 & 85.89 & 85.30 & 83.96 & 80.62 & 73.96 & 69.40 & 75.67 & 80.43 & 82.67 & 82.97 & 0.66 \\
\midrule
\rowcolor{black!4}
\multicolumn{13}{@{}l}{\textit{Pythia, WikiText PPL}} \\
\quad 14M & 138.64 & 156.44 & 240.76 & 320.79 & 352.91 & 353.30 & 329.13 & 271.16 & 192.75 & 144.78 & 165.88 & 0.86 \\
\quad 70M & 63.09 & 42.77 & 44.27 & 53.15 & 63.31 & 68.91 & 65.41 & 55.65 & 46.28 & 44.82 & 66.30 & 0.06 \\
\quad 160M & 71.27 & 45.44 & 50.47 & 65.79 & 82.47 & 73.33 & 73.70 & 73.04 & 54.11 & 47.81 & 64.67 & 0.18 \\
\quad 410M & 22.31 & 16.72 & 18.66 & 23.81 & 30.25 & 33.76 & 30.59 & 24.32 & 19.24 & 17.45 & 24.46 & 0.37 \\
\quad 1B & 17.69 & 12.99 & 13.73 & 16.15 & 19.76 & 22.25 & 20.11 & 16.47 & 13.91 & 13.12 & 17.64 & 0.23 \\
\quad 1.4B & 15.91 & 11.84 & 12.32 & 14.27 & 17.47 & 20.43 & 19.38 & 15.84 & 13.15 & 12.21 & 18.10 & 0.19 \\
\quad 6.9B & 13.22 & 9.72 & 9.84 & 11.11 & 13.42 & 15.71 & 14.79 & 12.09 & 10.28 & 9.80 & 12.98 & 0.18 \\
\midrule
\rowcolor{black!4}
\multicolumn{13}{@{}l}{\textit{Other Models, WikiText PPL}} \\
\quad HF 1.8B & 15.71 & 11.73 & 12.26 & 14.90 & 20.02 & 26.72 & 24.02 & 17.16 & 13.15 & 11.86 & 15.01 & 0.55 \\
\quad OLMo-7B & 10.83 & 9.84 & 12.60 & 20.68 & 42.19 & 51.80 & 25.16 & 15.41 & 11.32 & 10.09 & 11.61 & 1.53 \\

\bottomrule
\end{tabularx}
\end{table}

\begin{table}[h]
\centering
\caption{Raw interpolation paths between checkpoints before matching. Large perplexities are written in scientific notation.}
\label{tab:raw-interpolation-paths}
\footnotesize
\setlength{\tabcolsep}{1.8pt}
\renewcommand{\arraystretch}{1.08}
\begin{tabularx}{\textwidth}{@{}>{\raggedright\arraybackslash}p{0.14\textwidth}*{11}{>{\centering\arraybackslash}X}@{\hspace{4pt}}|@{\hspace{4pt}}>{\centering\arraybackslash}X@{}}
\toprule
\multirow{2}{*}{Model} & \multicolumn{11}{c}{Interpolation coefficient $\lambda$} & \multirow{2}{*}{\shortstack{Loss\\Barrier}} \\
\cmidrule(lr){2-12}
& $0.0$ & $0.1$ & $0.2$ & $0.3$ & $0.4$ & $0.5$ & $0.6$ & $0.7$ & $0.8$ & $0.9$ & $1.0$ & \\
\midrule
\rowcolor{black!4}
\multicolumn{13}{@{}l}{\textit{ViT, Accuracy (\%)}} \\
\quad Small & 81.46 & 75.88 & 32.43 & 0.89 & 0.16 & 0.13 & 0.12 & 0.27 & 6.08 & 62.60 & 78.71 & 6.58 \\
\quad Base & 84.58 & 75.93 & 0.45 & 0.11 & 0.10 & 0.10 & 0.22 & 2.90 & 54.28 & 76.60 & 79.14 & 6.57 \\
\quad Large & 85.82 & 84.25 & 65.73 & 0.46 & 0.09 & 0.10 & 0.10 & 0.06 & 6.36 & 69.28 & 83.00 & 6.47 \\
\midrule
\rowcolor{black!4}
\multicolumn{13}{@{}l}{\textit{Pythia, WikiText PPL}} \\
\quad 14M & 138.35 & 4.49e3 & 6.02e4 & 3.56e5 & 1.77e6 & 2.97e6 & 1.31e6 & 2.42e5 & 1.12e4 & 638.47 & 165.04 & 9.89 \\
\quad 70M & 63.05 & 188.16 & 1.09e3 & 5.44e3 & 1.67e4 & 2.36e4 & 1.45e4 & 4.37e3 & 849.15 & 134.67 & 66.16 & 5.90 \\
\quad 160M & 71.19 & 140.68 & 678.87 & 4.15e3 & 2.02e4 & 3.61e4 & 1.93e4 & 5.17e3 & 899.95 & 154.55 & 64.09 & 6.28 \\
\quad 410M & 22.35 & 45.10 & 161.37 & 643.52 & 2.11e3 & 3.45e3 & 2.28e3 & 677.88 & 153.22 & 48.98 & 24.50 & 4.99 \\
\quad 1B & 17.69 & 23.82 & 76.79 & 367.14 & 1.17e3 & 1.76e3 & 1.11e3 & 348.00 & 72.13 & 23.49 & 17.63 & 4.60 \\
\quad 1.4B & 15.95 & 22.92 & 78.35 & 603.05 & 3.37e3 & 6.31e3 & 3.82e3 & 707.46 & 98.97 & 28.56 & 18.40 & 5.91 \\
\quad 6.9B & 13.43 & 16.21 & 29.49 & 92.80 & 321.88 & 557.83 & 303.33 & 87.71 & 26.29 & 15.03 & 13.03 & 3.74 \\
\midrule
\rowcolor{black!4}
\multicolumn{13}{@{}l}{\textit{Other Models, WikiText PPL}} \\
\quad HF 1.8B & 15.71 & 19.22 & 116.63 & 8.62e3 & 2.07e4 & 2.79e4 & 2.40e4 & 1.26e4 & 177.02 & 18.55 & 15.01 & 7.50 \\
\quad OLMo-7B & 10.80 & 13.63 & 153.74 & 5.38e3 & 3.64e4 & 3.06e4 & 5.66e3 & 734.08 & 34.20 & 12.95 & 11.45 & 8.10 \\

\bottomrule
\end{tabularx}
\end{table}

\begin{table}[t]
\centering
\caption{Zero-shot commonsense accuracy for Pythia endpoints and matched models.}
\label{tab:pythia_common_tasks}
\begin{tabular}{llccc}
\toprule
Model & Method & PIQA & SciQ & ARC-E \\
\midrule
410M & Original Pythia & 66.43 & 81.50 & 51.89 \\
410M & Raw Interpolation & 55.39 & 31.50 & 29.71 \\
410M & Learned Matching & 55.44 & 56.30 & 33.25 \\
410M & Dual Learned Matching & 55.71 & 60.00 & 35.69 \\
\midrule
1B & Original Endpoint & 70.95 & 83.90 & 56.73 \\
1B & Raw Interpolation & 54.35 & 35.50 & 29.08 \\
1B & Learned Matching & 55.98 & 64.80 & 36.57 \\
1B & Dual Learned Matching & 56.42 & 64.40 & 37.16 \\
\bottomrule
\end{tabular}
\end{table}

Table~\ref{tab:pythia_common_tasks} gives a preliminary downstream check on zero-shot commonsense tasks. Although LMC-DM improves over raw interpolation, the matched interpolated models still lag behind the original endpoints, indicating that low WikiText barriers do not yet imply lossless downstream behavior. This is a limitation of the current setup: Pythia is a base model, and due to limited resources, our symmetry learning uses only hundreds of millions of tokens without downstream supervision; scaling this training budget may improve downstream preservation, which we leave to future work.

\section{Normalization Reparameterization Details}
\label{app:norm-details}

This appendix gives the algebra behind the normalization reparameterization used in
Section~\ref{sec:pre}. We follow the same row-vector convention as in the main text. Let
\[
\mathbf{C}=\mathbf{I}-\frac{1}{d}\mathbf{1}\mathbf{1}^{\top}
\]
be the channel-centering operator. Writing $\mathrm{LayerNorm}_0$ and
$\mathrm{RMSNorm}_0$ for the corresponding parameter-free normalization
operators, the non-affine part of LayerNorm can be written as RMSNorm on the
centered residual stream:
\[
\mathrm{LayerNorm}_0(\mathbf{X})
=
\mathrm{RMSNorm}_0(\mathbf{X}\mathbf{C}).
\]
Thus a pre-LayerNorm Transformer can be represented in a parameter-free
pre-RMSNorm form by keeping residual writes centered. If a branch writes to the
residual stream as
\[
\mathbf{Y}
=
\mathbf{U}\mathbf{W}_{\mathrm{out}}
+
\mathbf{1}\mathbf{b}_{\mathrm{out}}^\top ,
\]
then its centered form is
\[
\mathbf{Y}\mathbf{C}
=
\mathbf{U}(\mathbf{W}_{\mathrm{out}}\mathbf{C})
+
\mathbf{1}(\mathbf{C}\mathbf{b}_{\mathrm{out}})^\top .
\]
Therefore the final weight matrix of each residual-writing branch is projected
as $\mathbf{W}_{\mathrm{out}}\mathbf{C}$. If a bias is present, only its
all-ones component is removed, while the centered component is kept or absorbed
into an adjacent affine term.

The affine parameters of LayerNorm or RMSNorm can also be absorbed into the next
linear layer. Let $\boldsymbol{\gamma}$ and $\boldsymbol{\beta}$ be the
normalization gain and bias, and let the following affine map have weight
$\mathbf{W}$ and bias $\mathbf{b}$. Then
\[
\begin{aligned}
\mathrm{LayerNorm}_{\boldsymbol{\gamma},\boldsymbol{\beta}}(\mathbf{X})
\mathbf{W}
+
\mathbf{1}\mathbf{b}^{\top}
&=
\mathrm{RMSNorm}_{0}(\mathbf{X}\mathbf{C})
\mathrm{Diag}(\boldsymbol{\gamma})\mathbf{W}
+
\mathbf{1}(\mathbf{b}+\mathbf{W}^{\top}\boldsymbol{\beta})^{\top}\\
&=
\mathrm{RMSNorm}_{0}(\mathbf{X}\mathbf{C})\mathbf{W}'
+
\mathbf{1}{\mathbf{b}'}^{\top},
\end{aligned}
\]
where
\[
\mathbf{W}'=\mathrm{Diag}(\boldsymbol{\gamma})\mathbf{W},
\qquad
\mathbf{b}'=\mathbf{b}+\mathbf{W}^{\top}\boldsymbol{\beta}.
\]

\section{Continuous Symmetry Parameterizations}

This section gives additional details on the continuous symmetry parameterizations used in LMC-DM. In all cases, the goal is not merely to introduce learnable variables, but to keep the transformed weights inside the valid functionality-preserving symmetry family throughout gradient-based optimization. In our implementation, both endpoint models maintain their own symmetry modules, initialized from weight matching when available.

\subsection{Cayley Transform with Signs}
For orthogonal symmetries, we mainly use the Cayley transform with a fixed sign component. Given a free matrix $\mathbf{S}$, we form a skew-symmetric matrix
\[
\mathbf{A}=\mathbf{S}-\mathbf{S}^{\top},
\]
and parameterize the smooth orthogonal factor as
\[
\mathbf{C}(\mathbf{A})=(\mathbf{I}-\mathbf{A})^{-1}(\mathbf{I}+\mathbf{A}).
\]
This guarantees $\mathbf{C}(\mathbf{A})^{\top}\mathbf{C}(\mathbf{A})=\mathbf{I}$ at every optimization step, avoiding the need for projection after each update. In the implementation, the signed variant composes this standard Cayley factor with a fixed signed left factor chosen at initialization, which allows us to represent weight matching solutions with positive or negative determinant without changing the optimization formula.

\subsection{Matrix Exponential Parameterization}
We use two matrix-exponential parameterizations, depending on whether the symmetry is constrained to be orthogonal or merely invertible. For orthogonal symmetries, the exponent is skew-symmetric:
\[
\mathbf{A}=\mathbf{S}-\mathbf{S}^{\top},\qquad
\mathbf{Q}=\exp(\mathbf{A}),\qquad
\mathbf{Q}^{-1}=\mathbf{Q}^{\top},
\]
with an optional fixed reflector handled in the implementation to match the determinant component selected at initialization. For general invertible symmetries, the exponent is unconstrained:
\[
\mathbf{M}=\exp(\mathbf{U}),\qquad
\mathbf{M}^{-1}=\exp(-\mathbf{U}),
\]
where $\mathbf{U}$ is a free matrix. The implementation again optionally composes this standard form with a fixed reflector when needed. This keeps $\mathbf{M}$ invertible by construction and gives an exact inverse.

\subsection{Polar-style Parameterization}
For invertible symmetries, polar decomposition style parameterization provides a promising alternative to the matrix exponential parameterization. We decompose the symmetry matrix into an orthogonal factor and a positive-definite factor,
\[
\mathbf{M}=\mathbf{Q}\mathbf{P},
\]
where $\mathbf{Q}$ is an orthogonal factor parameterized by the Cayley transform previously discussed, and $\mathbf{P}$ is positive definite. The positive-definite factor $\mathbf{P}$ is represented as
\[
\mathbf{P}=\mathbf{L}\mathbf{L}^{\top},
\]
where $\mathbf{L}$ is lower triangular with positive diagonal entries. This parameterization keeps the matrix invertible, provides a stable inverse
\[
\mathbf{M}^{-1}=\mathbf{P}^{-1}\mathbf{Q}^{\top},
\]
and separates rotational and scaling/shearing degrees of freedom, which we found more stable than unconstrained direct matrices in some settings.

\subsection{Potential Alternative Parameterizations}
We also considered several other alternatives. Direct invertible parameterization optimizes a free matrix and computes its inverse explicitly; it is simple and expressive, but can become ill-conditioned during training. 
For orthogonal symmetry, SVD projection parameterizes an unconstrained matrix and projects it to the nearest orthogonal factor in the forward pass;  while flexible, this introduces an expensive projection and can lead to instability in backpropagation. Householder products provide a structured orthogonal parameterization, but we found them relatively slow on large matrices in GPU-based training.

\section{Further Discussion}
\label{app:further_discussion}

\noindent \textbf{Insights from Linear Mode Connectivity.}
Our study demonstrates that independently trained checkpoints exhibit linear mode connectivity when accounting for weight-space symmetries. This suggests that neural networks converge to functionally similar solutions within a structured, low-loss region of the parameter space, rather than isolated minima. These findings align with the Platonic Representation Hypothesis~\cite{huh2024platonic}, suggesting that representational convergence extends into the weight space itself. This perspective reveals a significant degree of structural redundancy that is often underestimated when models are viewed only through their raw parameterization.

\noindent \textbf{Potential Applications.} We provide several potential applications of the proposed LMC-DM method:
\begin{itemize}
    \item \textbf{Weight Space Learning and Generation:}
    By resolving symmetries between different checkpoints, LMC-DM produces a family of aligned models that share a consistent weight-space structure. This may facilitate training generative models over neural network weights. In addition, low-loss interpolation paths themselves provide a principled way to sample new models, which could be useful for model ensembling or as training data for weight-space generative methods.

    \item \textbf{Federated Learning:}
    Symmetry-based alignment can reduce inter-client variance by mapping locally trained models into a shared coordinate system, potentially improving aggregation quality and convergence stability in federated learning.

    \item \textbf{Efficient AI:}
    If independently trained models can be reliably merged without performance degradation, it becomes possible to consolidate capabilities from multiple checkpoints without full retraining. This could significantly reduce computational cost, energy consumption, and barriers to large-scale model development.
\end{itemize}


\end{document}